\documentclass[runningheads]{llncs}
\usepackage{graphicx}
\usepackage{amsmath,amssymb} %
\usepackage{color}
\usepackage[width=122mm,left=12mm,paperwidth=146mm,height=193mm,top=12mm,paperheight=217mm]{geometry}

\usepackage{times}
\usepackage{epsfig}
\usepackage{comment}
\usepackage{multirow}
\usepackage[compress, nosort]{cite}

\usepackage{sidecap}
\usepackage[font=small,labelfont=bf]{caption}

\def\bx{ {\bf x } }

\newcommand\CFM[1][]{{$\tt CFM#1$}}

\newcommand{\etal}{\mbox{\emph{et al.\ }}}

\begin{document}
\pagestyle{headings}
\mainmatter

\title{Pushing the Limits of Deep CNNs for Pedestrian Detection\thanks{Corresponding author: C. Shen (e-mail: chunhua.shen@adelaide.edu.au). }
} %

\titlerunning{Pushing the Limits of Deep CNNs for Pedestrian Detection}

\authorrunning{Hu, Wang, Shen, van den Hengel, and Porikli}

\author{Qichang Hu$^1$, Peng Wang$^1$, Chunhua Shen$^1$, Anton van den Hengel$^1$, Fatih Porikli$^2$ }
\institute{$^1$University of Adelaide, ~ ~$^2$NICTA}

\maketitle

\begin{abstract}
   Compared to other applications in computer vision, convolutional neural networks have under-performed on pedestrian detection.
   A breakthrough was made very recently by using sophisticated deep CNN models, with a number of hand-crafted features
   \cite{cai2015learning}, or explicit occlusion handling mechanism \cite{tiandeep}.
   In this work, we show that by re-using the convolutional feature maps (CFMs) of a deep convolutional neural network (DCNN) model
   as image features to train an ensemble of boosted decision models,
   we are able to achieve the best reported accuracy without using specially designed learning algorithms.
   We empirically identify and disclose important implementation details. 
   We also show that pixel labelling may be simply combined with a detector to boost the detection performance.
   By adding complementary hand-crafted features such as optical flow, the DCNN based detector can be
   further improved. We set a new record on the Caltech pedestrian dataset, lowering 
   the log-average miss rate from $11.7\%$ to $8.9\%$, a relative improvement of $24\%$. We also 
   achieve a comparable result to the state-of-the-art approaches on the KITTI dataset.

\end{abstract}

\section{Introduction}

The problem of pedestrian detection has been intensively studied in recent
years. Prior to the very recent work in deep convolutional neural networks (DCNNs) based methods
\cite{cai2015learning,tiandeep}, the top performing pedestrian detectors are boosted decision forests with carefully
hand-crafted features, such as histogram of gradients (HOG)~\cite{dalal2005histograms}, self-similarity
(SS)~\cite{shechtman2007matching}, aggregate channel features (ACF)~\cite{dollar2014fast}, filtered channel
features~\cite{zhang2015filtered} and optical flow~\cite{paisitkriangkrai2014pedestrian}.

Recently, DCNNs have significantly outperformed comparable methods on a wide variety of vision problems~\cite{krizhevsky2012imagenet,simonyan2014very,
szegedy2014going,girshick2014rich,tompson2014joint,hariharan2014simultaneous,simonyan2014two,branson2014bird}.
A region-based convolutional neural network (R-CNN)~\cite{girshick2014rich} achieved excellent performance
for {\em generic} object detection, for example,
in which a set of potential detections (object proposals) are evaluated by a DCNN model.
CifarNet~\cite{krizhevsky2009learning} and AlexNet~\cite{krizhevsky2012imagenet}
have been extensively evaluated in the R-CNN detection framework in \cite{hosang2015taking} for pedestrian detection.
In their work, the best performance ($23.3\%$) was achieved by AlexNet pre-trained on the ImageNet~\cite{deng2009imagenet} classification dataset.
Note that this result is still inferior to conventional pedestrian detectors such as \cite{zhang2015filtered}
and \cite{paisitkriangkrai2014pedestrian}.
The DCNN models in \cite{hosang2015taking} under-perform mainly because the network design is not optimal for pedestrian detection. 
The performance of R-CNNs for pedestrian detection has further improved to
$16.43\%$ in \cite{tiandeep} through the use of a deeper GoogLeNet  model
which is fine-tuned using Caltech pedestrian dataset.

To explicitly model the deformation and occlusion,
another line of research for object detection is part-based models~\cite{enzweiler2010multi,felzenszwalb2010object,lin2015discriminatively,girshick2014deformable}
and explicit occlusion handling~\cite{mathias2013handling,ouyang2013single,tang2014detection}.
DCNNs have also been incorporated along this stream of work for pedestrian detection~\cite{ouyang2012discriminative,ouyang2013joint,luo2014switchable},
but none of these approaches has achieved better results than the best hand-crafted features based method of \cite{zhang2015filtered} on the Caltech dataset.

The performance of pedestrian detection is improved over hand-crafted features by a large margin (a $\sim5\%$ gain on Caltech),
by two very recent approaches relying on DCNNs:
CompACT-Deep~\cite{cai2015learning} combines hand-crafted features and fine-tuned DCNNs into a complexity-aware cascade.
Tian \etal~\cite{tiandeep} fine-tuned a pool of part detectors using a pre-trained GoogLeNet, and
the resulting ensemble model (refer to as DeepParts) delivers similar results as CompACT-Deep. 
Both approaches are much more sophisticated than the standard R-CNN framework:
CompACT-Deep involves the use of a variety of hand-crafted features, a small CNN model and a large VGG$16$ model~\cite{simonyan2014very}. 
DeepParts contains $45$ fine-tuned DCNN models and
needs a set of strategies (including bounding-box shifting handling and part selection) to arrive at the reported result.
Note that the high complexity of DCNN models can lead to practical difficulties.
For example, it can be too costly to load all 45 DCNN models into a GPU card.

Here we ask a question:
Is a complex DCNN based learning approach really a must for achieving the state-of-the-art performance? 
Our answer to this question is negative.
In this work, we propose alternative methods for pedestrian detection,
which are simpler in design, with comparable or even better performance. Firstly, we extensively evaluate the CFMs extracted from 
convolutional layers of a fine-tuned VGG$16$ model for pedestrian detection.
Using only a CFM of a single convolutional layer, we train a boosted-tree-based detector and the resulting model already 
significantly outperforms all previous methods except the above two sophisticated DCNN frameworks.
This model can be seen as a strong baseline for pedestrian detection as it is very simple in terms of implementation.

We show that the CFMs from multiple convolutional layers can be used for training effective boosted decision forests.
These boosted decision forests are combined altogether simply by score averaging.
The resulting ensemble model beats all competing methods on the Caltech dataset.
We further improve the detection performance by incorporating a semantic pixel labelling model.
Next we review some related work.

\subsection{Related Work}

\subsubsection{Convolutional feature maps (CFMs)}

It has been shown in \cite{ren2015object,hariharan2014hypercolumns,ccf2015Yang}
that CFMs have strong representation abilities for many tasks. Long
\etal~\cite{long2014fully} cast all fully-connected layers in DCNNs as
convolutions for semantic image segmentation. In
\cite{hariharan2014hypercolumns}, the CFMs from multiple layers are stacked
into one vector and used for segmentation and localization. Ren
\etal~\cite{ren2015object} learn a network on the CFMs (pooled to a fixed
size) of a pre-trained model.

The work by Yang \etal~\cite{ccf2015Yang} is close to ours,
which trains a boosted decision forest for pedestrian detection with the CFM features from
the Conv$3$-$3$ layer of the VGG$16$ model~\cite{simonyan2014very},
and the performance ($17.32\%$) on Caltech is comparable to checkerboards~\cite{zhang2015filtered}.
It seems that there is no significant superiority of the CFM used in \cite{ccf2015Yang}
over hand-crafted features on the task of pedestrian detection.
The reason may be two-fold. First, the CFM used in \cite{ccf2015Yang} are extract from the pre-trained
VGG$16$ model which is {\em not fine-tuned on a pedestrian dataset};
Second, CFM features are extracted from only one layer and the multi-layer structure of DCNNs is not fully exploited. 
We show in this work that both of these two issues are critically important in achieving good performance.

\subsubsection{Segmentation for object detection}

The cues used by segmentation approaches are typically complementary to those exploited by top-down methods.
Recently, Yan \etal~\cite{yan2015object} propose
to perform generic object detection by labelling super-pixels,
which results in an energy minimization problem with data term learned by DCNN models.
In \cite{fidler2013bottom,hariharan2014simultaneous}, segmented image regions (not bounding boxes) are generated as object 
proposals and then used for object detection.

In contrast to the above region (or super-pixel) based methods, we here exploit at an even finer level of information,
that is, pixel labelling. In particular, in this work we demonstrate that we can improve the detection performance by simply re-scoring the proposals generated by a detector, using pixel-level scores.

\subsection{Contributions}

We revisit pedestrian detection with DCNNs by studying the impact of a few training details and design parameters.
We show that fine-tuning of a DCNN model using pedestrian data is critically important. Proper bootstrapping has a considerable impact too. Besides these findings, other main contributions of this work can be summarized as follows.
\begin{enumerate}
  \itemsep +2pt%

\item
{\em The use of multi-layer CFMs for training a state-of-the-art pedestrian detector.} 
We show that it is possible to train an 
ensemble of boosted decision forests using multi-layer CFMs that 
outperform all previous methods. For example, with CFM features extracted from two convolutional layers, we can achieve a log-average miss rate of 
$ 10.7\% $ on Caltech, which already perform better than all previous methods, including the two sophisticated DCNNs based methods 
\cite{cai2015learning,tiandeep}. %

\item
{\em Incorporating semantic pixel labelling.}
We also propose a combination of sliding-window detectors and semantic pixel-labelling, which performs on par with the best of previous methods. 
To keep the method simple, we use the weighted sum of pixel-labelling scores within a proposal region to represent the score of the proposal.

\item
{\em The best reported pedestrian detector.}
A new performance record for Caltech is set by exploiting a DCNN as well as two complimentary hand-crafted features: ACF and optical-flow features. 
This shows that some types of hand-crafted features are complementary to deep convolutional features.%

\end{enumerate}

Before we present our methods, we briefly describe the datasets, evaluation metric and boosting models in our experiments.

\subsection{Datasets, Evaluation metric and Models}

\noindent \textbf{Caltech pedestrian dataset} The Caltech dataset~\cite{dollar2012pedestrian} is
one of the most popular datasets for pedestrian detection. It contains $250$k frames extracted from $10$ hours of urban traffic video. There are in total $350$k annotated bounding boxes with $2300$ unique pedestrians.
The standard training set and test set consider one out of each $30$ frames.
In our experiments, the training images are increased to one out of each $4$ frames.
Note that many competing methods~\cite{zhang2015filtered,ccf2015Yang,hosang2015taking} have used the same extended training set or
even more data (every third frame). %

For Caltech dataset, we evaluate the performance of various detectors using the log-average miss rate (MR)
which is computed by averaging the miss rate at false positive rates spaced evenly between $0.01$ to $1$ false-positives-per-image 
(FPPI) range. Unless otherwise specified, the detection performance on our experiments shown in the remainder of the paper is the MR on the Caltech $\sf{Reasonable}$ test set.

\noindent \textbf{KITTI pedestrian dataset} The KITTI dataset~\cite{geiger2012kitti} consists of 7481 training images and 7518 test images, comprising more than 80 thousands of annotated objects in traffic scenes. The KITTI dataset provides a large number of pedestrians with different sizes, viewpoints, occlusions, and truncations. Due to the diversity of these objects, the dataset has three subsets ($\sf{Easy}$, $\sf{Moderate}$, $\sf{Hard}$) with respect to the difficulty of object size, occlusion and truncation. We use the $\sf{Moderate}$ training subset as the training data in our experiments.

For KITTI dataset, average precision (AP) is used to evaluate the detection performance. The average precision summaries the shape of the precision-recall curve, and is defined as the mean precision at a set of evenly spaced recall levels. All methods are ranked based on the $\sf{Moderate}$ difficult results. %

\noindent \textbf{Boosted decision forest}
Unless otherwise specified, we train all our boosted decision forests using the following parameters.
The boosted decision model consists of $4096$ depth-$5$ decision trees, %
trained via the shrinkage version of real-Adaboost~\cite{hastie2005elements}.
The size of model is set to $128\times64$ pixels, and one bootstrapping iteration is implemented 
to collect hard negatives and re-trains the model.
The sliding window stride is set to $4$ pixels. %

\section{Boosted Decision Forests with Multi-layer CFMs}
\label{sec:approach1}

In this section, we firstly show that the performance of boosted decision forests with CFMs can be significantly improved
by simply fine-tuning DCNNs with hard negative data extracted through bootstrapping.
Then boosted decision forests are trained with different layers of CFMs,
and the resulting ensemble model is able to achieve the best reported result on the Caltech dataset.

\subsection{Fine-tuning DCNNs with Bootstrapped Data}

In this work, the VGG$16$~\cite{simonyan2014very} model is used to extract CFMs.
As we know, the VGG$16$ model was originally pre-trained on the ImageNet data
with image-level annotations and was not trained specifically for the pedestrian detection task.
It is expected that the detection performance of boosted decision forests trained with CFMs
ought to be improved by fine-tuning the VGG$16$ model with Caltech pedestrian data.

To adapt the pre-trained VGG$16$ model to the pedestrian detection task, we modify the structure of the model. 
We replace the $1000$-way classification layer with a randomly initialized binary classification layer and
change the input size from $224\times224$ to $128\times64$ pixels.
We also reduce the number of neurons in fully connected layers from $4096$ to $2048$.
We fine-tune all layers of this modified VGG$16$, except the first $4$ convolutional layers since they correspond to low-level 
features which are largely universal for most visual objects.
The initial learning rate is set to $0.001$ for convolutional layers and $0.01$ for fully connected layers.
The learning rate is divided by $10$ at every $10000$ iterations.
For fine-tuning, $30$k positive and $90$k negative examples are collected by different approaches.
The positive samples are those overlapping with a ground-truth bounding box by $[0.5, 1]$,
and the negative samples by $[0, 0.25]$.
At each SGD iteration, we uniformly sample $32$ positive samples and $96$ negative samples to construct a mini-batch of size $128$.

We train boosted decision forests with the CFM extracted from the Conv$3$-$3$ layer
of differently fine-tuned VGG$16$ models and the results are shown in Table~\ref{tab:fine-tuning}.
Note that all the VGG$16$ models in this table are fine-tuned from the original model pre-trained on ImageNet data.
It can be observed that the log-average miss rate is reduced from $18.71\%$ to $16.42\%$
by replacing the pre-trained VGG$16$ model with the one fine-tuned on data collected by applying an ACF~\cite{dollar2014fast} 
detector on the training dataset.
The detection performance is further improved to $14.54\%$ MR if it is fine-tuned on the bootstrapped data
using the previous trained model \CFM3b.
Another $1\%$ performance gain is obtained by applying shrinkage to the coefficients of weak learners,
with shrinkage parameter being $0.5$ (see \cite{paisitkriangkrai2014strengthening}).
The last model (corresponding to row $4$ in Table~\ref{tab:fine-tuning}) is referred to as \CFM3 from now on.

\begin{table}[tbp!]
\begin{center}
\caption{Performance improvements with different fine-tuning strategies and shrinkage (on $\sf{Reasonable}$).
         All boosted decision forests are trained with the CFM extracted from the Conv$3$-$3$ layer of VGG$16$.
         \CFM3a: the original VGG$16$ model pre-trained on ImageNet is used to extract features.
         \CFM3b: the VGG$16$ model is fine-tuned with the data collected by an ACF~\cite{dollar2014fast} detector.
         \CFM3c and \CFM3: the fine-tuning data is obtained by bootstrapping with \CFM3b.
         With the same fine-tuning data, setting the shrinkage parameter of Adaboost to $0.5$
          brings an additional $1\%$ reduction on the MR}
\label{tab:fine-tuning}
\begin{tabular}{cccc}
\hline
Model & Fine-tuning data   & Shrinkage & Miss rate (\%) \\
\hline\hline
\CFM3a  & No fine-tuning             & $-$   & $18.71$ \\
\CFM3b  & Collected by ACF           & $-$   & $16.42$ \\
\CFM3c  & Bootstrapping with \CFM3b  & $-$   & $14.54$ \\
\CFM3   & Bootstrapping with \CFM3b  & $0.5$ & $\textbf{13.49}$ \\
\hline
\end{tabular}
\end{center}
\vspace{-15pt}
\end{table}

\subsection{Ensemble of Boosted Decision Forests}

In the last experiment, we only use a CFM from a single layer of the VGG$16$ model. 
In this section, we intensively explore the deep structure of the
VGG$16$ model which consists of $13$ convolutional layers, $2$ fully connected
layers, and $1$ classification layer. These $13$ convolutional layers are
organized into $5$ convolutional stacks, convolutional layers in the same stack
have the same down-sampling ratio. We ignore the CFMs of the first two
convolutional stacks (each one contains $2$ layers) since they are universal for most visual objects. 

We train boosted decision forests with CFMs from individual convolutional layers of the VGG$16$ model which is the one fine-tuned with bootstrapped data (same as row $4$ in Table~\ref{tab:fine-tuning}).
All boosted decision forests are trained with the same data as \CFM3.
For models with Conv$3$-$\bx$ features, 
the input image are directly applied on the convolutional layers 
and resulting in a feature map with the down-sampling ratio of $4$.
The corresponding boosted decision forests work as a sliding-window detector with step-size of $4$.
For models with Conv$4$-$\bx$ and Conv$5$-$\bx$ features, they are applied to proposals generated by \CFM3 model. 
This is due to the large downsampling ratio of Conv$4$-$\bx$ and Conv$5$-$\bx$.
If the step-size of the sliding-window detector is too large, it will hurt the detection performance.

\begin{table}[tbp!]
\begin{center}
\caption{Comparison of detection performance (on $\sf{Reasonable}$) of boosted decision forests trained on individual CFMs.
         Note that models with Conv$3$-$\bx$ features works as sliding-window detectors,
         and models with Conv$4$-$\bx$ and Conv$5$-$\bx$ features are applied to the proposals generated by \CFM3.
         The top performing layers in each convolutional stack are Conv$3$-$3$, Conv$4$-$3$ and Conv$5$-$1$ respectively.
         The models trained with these three layers are denoted as \CFM3, \CFM4, and \CFM5 respectively
}
\label{tab:comparison of CFMs}
\begin{tabular}{r|c c c}
\hline
& & & \\ [-2ex]
Convolutional & $\#$ Channels & Down-sampling & Miss rate (\%) \\
layer &  & ratio &  \\
\hline\hline
& & & \\ [-2ex]
Conv$3$-$1$ & $256$ & $4$ & $19.15$ \\
Conv$3$-$2$ & $256$ & $4$ & $16.25$ \\
Conv$3$-$3$ (\CFM3) & $256$ & $4$ & $\mathbf{13.49}$ \\
\hline
& & & \\ [-2ex]
Conv$4$-$1$ & $512$ & $8$ & $12.95$ \\
Conv$4$-$2$ & $512$ & $8$ & $12.68$ \\
Conv$4$-$3$ (\CFM4) & $512$ & $8$ & $\mathbf{12.21}$ \\
\hline
& & & \\ [-2ex]
Conv$5$-$1$ (\CFM5) & $512$ & $16$ & $\mathbf{14.17}$ \\
Conv$5$-$2$ & $512$ & $16$ & $14.56$ \\
Conv$5$-$3$ & $512$ & $16$ & $18.24$ \\
\hline
\end{tabular}
\end{center}
\vspace{-25pt}
\end{table}

\begin{SCfigure*}[][tbp!]
\centering
\includegraphics[width=0.425\textwidth]{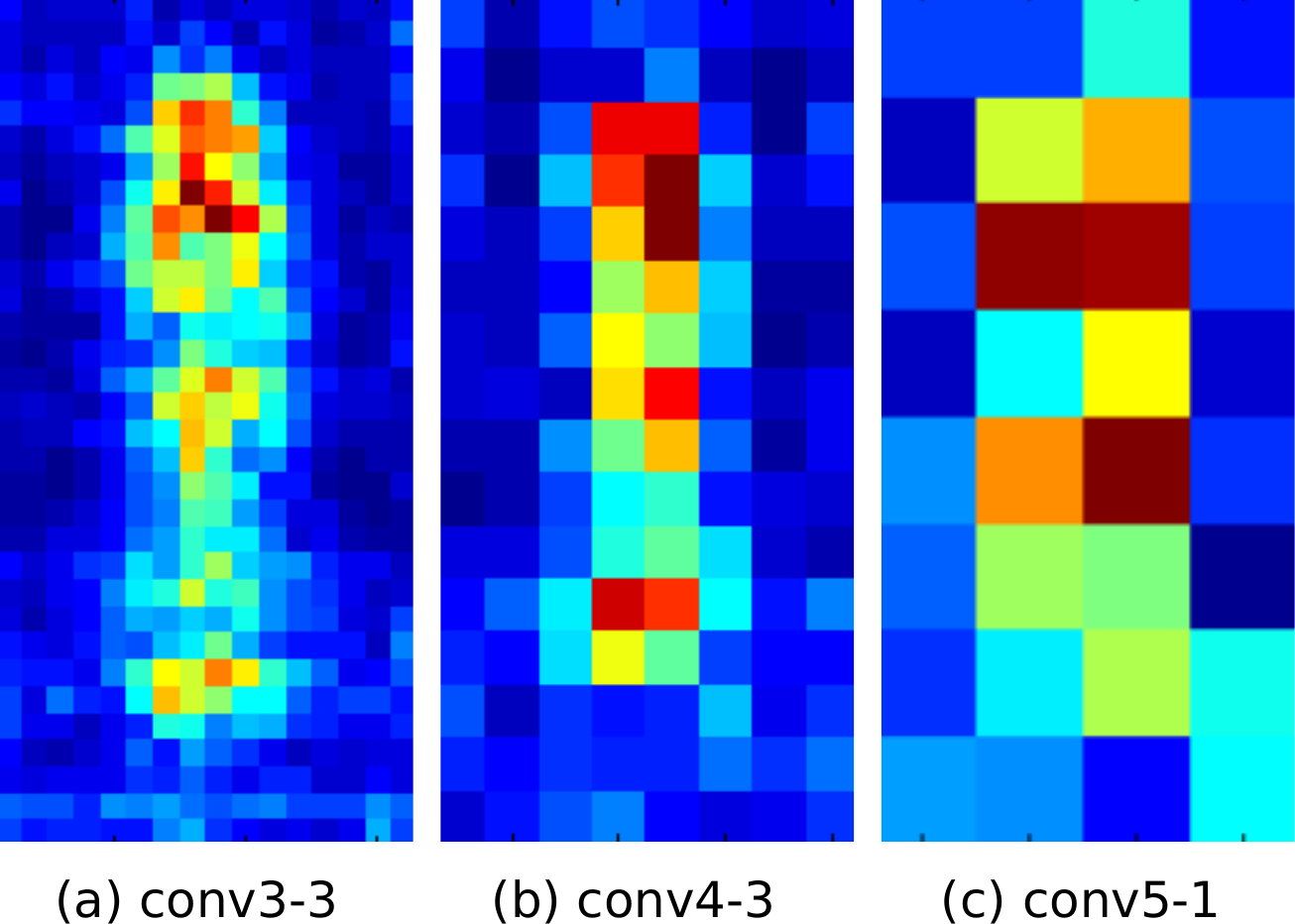}
\caption{The spatial distribution of CFMs selected by boosting algorithms.
         For a $128\times64$ input image, the size of feature maps are $32\times16$, $16\times8$, $8\times4$ respectively.
         Red pixels indicate that a large number of features are selected in that region and blue pixels correspond to low 
         frequency regions. The most important region correspond to the head, shoulder, waist and feet of a human.}
\label{fig:heatmaps}
\vspace{-5pt}
\end{SCfigure*}

Table~\ref{tab:comparison of CFMs} shows the comparison of detection performance of
these boosted decision forests on Caltech $\sf{Reasonable}$ setting.
We can observe that the MR is relatively high for the Conv$3$-$1$ layer and the Conv$5$-$3$ layer. 
We conjecture that the Conv$3$-$1$ layer provides relatively low-level features
which result in an under-fitting training.
In contrast, the semantic information in the Conv$5$-$3$ layer
may be too coarse for pedestrian detection.
According to Table~\ref{tab:comparison of CFMs},
the best performing layer in each convolutional stack, are
from inner layers of Conv$3$-$3$ (\CFM3), Conv$4$-$3$ (\CFM4), and Conv$5$-$1$ (\CFM5) respectively.
Fig.~\ref{fig:heatmaps} shows the spatial distribution of convolutional features,
which are frequently selected by above three \CFM{} models.
We observe that most active regions correspond to important human-body parts (such as head and shoulder).

\begin{figure*}[tbp!]
\begin{center}
   \includegraphics[width=0.9\textwidth]{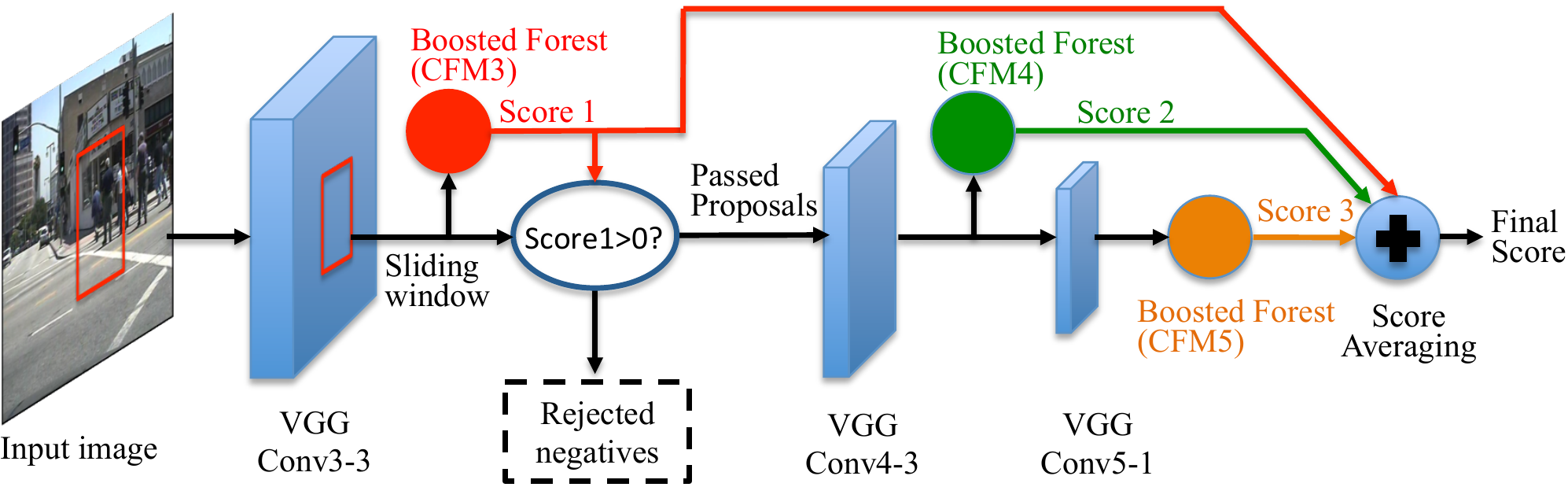}
\end{center}
\caption{The framework of an ensemble of boosted decision forests with multi-layer CFMs (\CFM3$+$\CFM4$+$\CFM5),
            which obtain a $10.46$ MR on the Caltech $\sf{Reasonable}$ test set.}
\label{fig:flow1}
\vspace{-5pt}
\end{figure*}

The boosted decision forests trained with CFMs of these three layers are further combined
together simply through score averaging.
Fig.~\ref{fig:flow1} shows the framework of the resulting ensemble model.
Firstly, \CFM3 model works as a sliding-window detector, which rejects the majority of negative examples and pass region proposals 
to \CFM4 and \CFM5. Both \CFM4 and \CFM5 generate the confidence score for each incoming proposal.
The final score is computed by averaging over the scores output by these three boosted decision forests.
{\em This model delivers the best reported log-average miss rate ($10.46\%$) on Caltech $\sf{Reasonable}$ setting without using any sophisticatedly designed algorithms.}

\begin{table}[tbp!]
\begin{center}
\caption{The comparison of performance (on $\sf{Reasonable}$) of different ensemble models.
DCNN: the entire VGG$16$ model fine-tuned by data collected by \CFM3.
The combination of multi-layer CFM models improves the detection performance of single-layer CFM models significantly ($3\%$)
}
\label{tab:ensemble-detectors}
\begin{tabular}{l c}
\hline
& \\ [-2ex]
Model combination & Avg. miss rate (\%) \\
\hline\hline
\CFM3$ + $\CFM4 & $10.68$ \\
\hline
\CFM3$ + $\CFM5 & $10.88$ \\
\hline
\CFM3$ + $\CFM4$ + $\CFM5 & $10.46$ \\
\hline
\CFM3$ + $\CFM4$ + $\CFM5$ + $DCNN & $\textbf{10.07}$ \\
\hline
\end{tabular}
\end{center}
\vspace{-20pt}
\end{table}

We also evaluate other combinations of the ensemble models.
Furthermore, a VGG$16$ model is fine-tuned with another round of bootstrapping (using \CFM3)
and its final output is also combined to improve the detection performance.
The corresponding results can be found in Table~\ref{tab:ensemble-detectors}
We can see that combining two layers already beats all existing approaches on Caltech,
and adding the entire large VGG$16$ model also gives a small improvement.

\section{Pixel Labelling Improves Pedestrian Detection}
\label{sec:approach2}

\begin{figure*}[tbp!]
\begin{center}
   \includegraphics[width=0.9\textwidth]{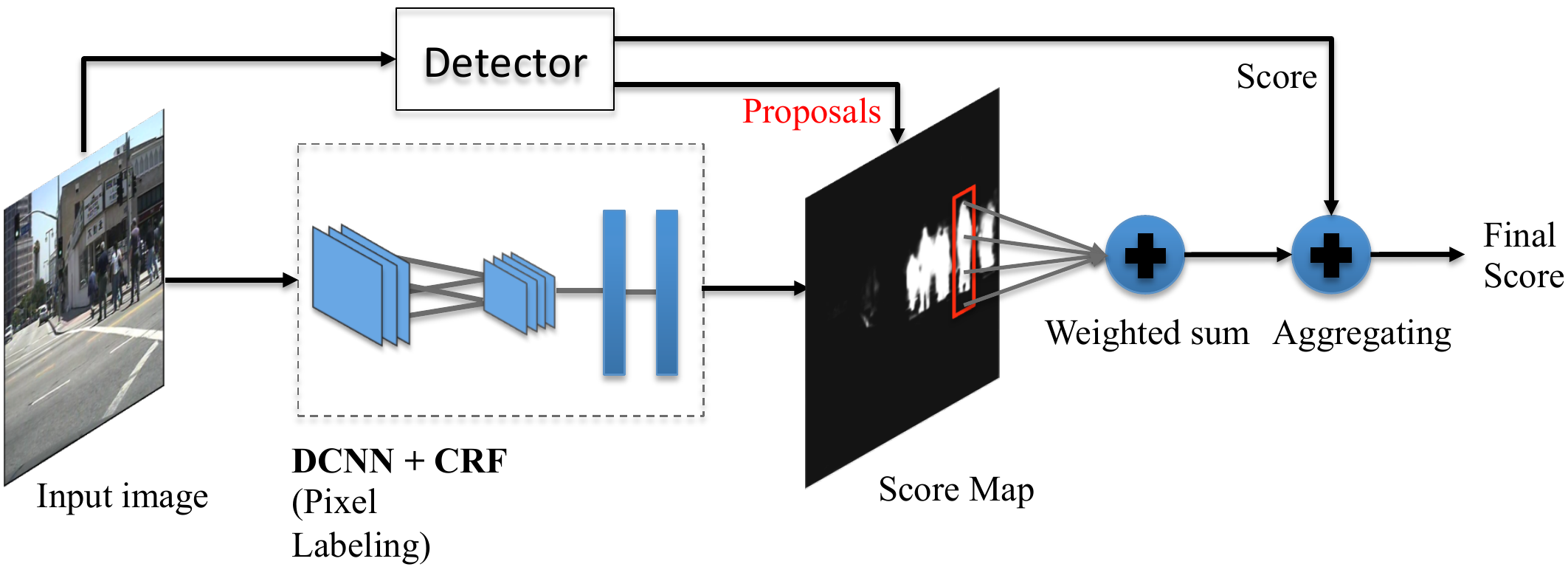}
\end{center}
   \caption{The framework for pedestrian detection with pixel-labelling.
            The region proposals and pixel-level score maps are obtained by
            individually applying the sliding-window detector and the pixel labelling model.
            Next, the weighted sum of pixel scores within a proposal region is aggregated with the detector
          score of the same proposal region. }
\label{fig:flow2}
\vspace{-15pt}
\end{figure*}

In this section, the sliding-window based detectors are enhanced by semantic pixel labelling.
By incorporating DCNNs, the performance of pixel labelling (semantic image segmentation) methods
have been recently improved 
significantly~\cite{long2014fully,chen2014semantic,hariharan2014hypercolumns,zheng2015conditional,lin2015efficient}.
In general, we argue that pixel labelling models encode information complementary to the sliding-window based detectors. 
Empirically, we show that consistent improvements are achieved over different types of detectors.

The segmentation method proposed in~\cite{chen2014semantic} is used here for pixel labelling,
in which a DCNN model (VGG$16$) is trained on the Cityscapes dataset~\cite{Cordts2015Cvprw}. 
The prediction map is refined by a fully-connected conditional random field (CRF)~\cite{koltun2011efficient}
with DCNN responses as unary terms. The Cityscapes dataset that we use for training is similar to the KITTI dataset which contains 
dense pixel annotations of 19 semantic classes such as road, building,
car, pedestrian, sky, etc. %
Note that our models that exploiting pixel labelling have used extra data for training on top of the Caltech dataset.
However, most deep learning based methods \cite{cai2015learning, tiandeep} have used extra data, at least the ImageNet dataset for 
pre-training the deep model. Pedestrian detection may benefit from the semantic pixel labelling in the following aspects:

$-$ {\em Multi-class information:} Learning from multiple classes, in contrast to the object detectors typically trained with two-class data,
the pixel labelling model carries richer object-level information.%

$-$ {\em Long-range context:} Using CRFs (especially fully-connected CRFs) as post-processing procedure,
many models (for example, \cite{chen2014semantic,lin2015efficient,zheng2015conditional}) have the ability to capture long-range context information. 
In contrast, sliding-window detectors only extract features from fixed-sized bounding boxes.

$-$ {\em Object parts:} The trained pixel labelling model may cater for more fine-grained details,
such that they are more insensitive to deformation and occlusion to some extent.

However, it is not straightforward to apply pixel labelling models to pedestrian detection problems.
One of the main impediments is that it is difficult to estimate the object bounding boxes from the pixel score map,
especially for people in crowds.

To this end, we propose to bring the pedestrian detector and pixel labelling model together.
In our framework (see Fig.~\ref{fig:flow2}), a sliding-window detector is responsible for providing region proposals and
a pixel labelling model is applied to the input image at the same time to generate a score map for the ``person'' class.
Next, a weighted mask is applied to the proposal region of the ``person'' score map to generate the weighted sum of pixel 
scores. Finally, the weighted sum and the detector score for the same proposal are aggregated together as the final score. 
The weighted mask is learned by averaging the pixel scores of ground truth region on the training images. To match the mask and
the input proposals, we resize both ground truth and test proposals to $100 \times 41$ pixels (no surrounding pixels).
Note that, there are more sophisticated methods for exploiting the labelling scores. For example, one can use the pixel labelling 
scores as the image features, similar to `object bank' \cite{LiOBank}, and train a linear model. 
In this work, we show that even simply weighted sum of the pixel scores considerably improves the results.

\begin{table}[tbp!]
\begin{center}
\caption{Performance improvements by aggregating pixel labelling models with sliding-window detectors
  		(on $ \sf Reasonable$).
         All the three detectors achieve performance gains,
         which shows that pixel labelling can be used to help detection.
         Note that the performance of our model `\CFM3 with Pixel labelling' is already {\em on par}
         with the previously best reported result of \cite{cai2015learning}
}
\label{tab:pixel-labelling}
\begin{tabular}{lcc}
\hline
Method & Avg.\ miss rate (\%) & Improve.  (\%)\\
\hline\hline
ACF   \cite{dollar2014fast}  & $22.23$ &\\
ACF$ + $Pixel label. & $17.73$ & $4.50$\\
\hline
Checkerboards  \cite{zhang2015filtered}  &  $18.25$ & \\
Checkerboards$ + $Pixel label. & $14.64$ &  $3.61$\\
\hline
\CFM3 (ours)& $13.49$ & \\
\CFM3$ + $Pixel label. & $11.58$ & $1.91$\\
\hline
\end{tabular}
\end{center}
\vspace{-15pt}
\end{table}

\begin{figure}[tbp!]
\centering
  \includegraphics[width=\textwidth]{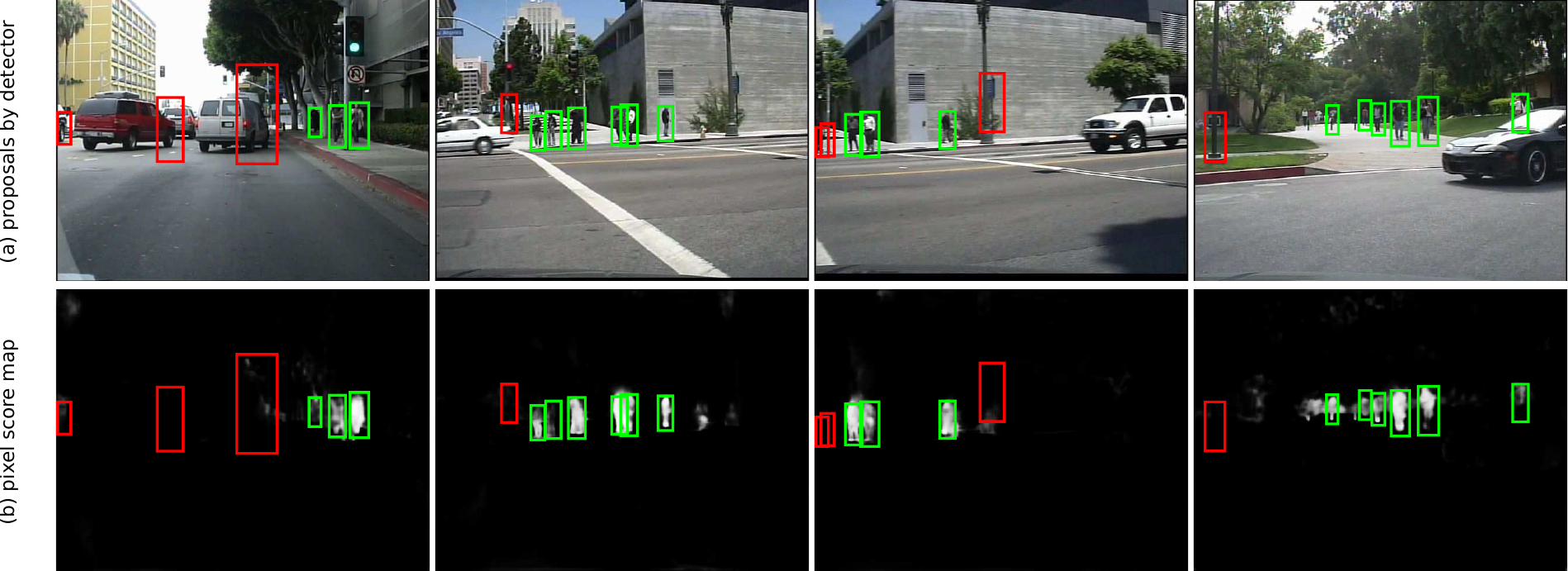}
\vspace{5pt}
\caption{Examples of some region proposals on the original images and the corresponding pixel score maps. %
A strong complementary relationship can be found in the generated proposals and the pixel score maps.
}
\label{fig:ped_seg}
\vspace{-20pt}
\end{figure}

Table~\ref{tab:pixel-labelling} shows the detection performance of different sliding-window detectors enhanced by pixel labelling.
Boosted decision forests are trained here with three types of features, which are ACF~\cite{dollar2014fast},
checkerboard features~\cite{zhang2015filtered} and the CFM from the Conv$3$-$3$ layer of VGG$16$ model (\CFM3).
We can see that the performances of all the three detectors are improved
by aggregating pixel labelling models. Fig.~\ref{fig:ped_seg} presents some region proposals on the original images and the 
corresponding pixel score maps.
Some of the false proposals generated by pedestrian detectors (\CFM3) can be removed by considering the context of a larger region 
(the largest bounding box in the first column in Fig.~\ref{fig:ped_seg}). Some occluded pedestrians have responses on the pixel 
score map (the rightmost bounding box in the fourth column in Fig.~\ref{fig:ped_seg}). This clearly illustrates why this 
combination works.

\section{Fusing Models}
\label{sec:approach3}

\subsection{Using Complementary Hand-crafted Features}

The detection performance of the \CFM3 model is critical in the proposed ensemble model,
since later components often reply on the detection results of this model.
In order to enhance the detection performance of the \CFM3 model, we make two variants of it by
combining two hand-crafted features: the ACF and optical flow.
We augment the \CFM3 features with the ACF and optical flow features to train an ensemble of
boosted decision forests. Optical flow features are extracted the same way as in \cite{paisitkriangkrai2014pedestrian}.

\begin{table}[tbp!]
\begin{center}
\caption{Comparison of detection results of different variants of the \CFM3 detector
(on $\sf Reasonable$).
The convolutional features of the Conv$3$-$3$ layer are combined with different types of hand-crafted features,
and used to train a boosted decision forest.
Both the performance of the variants and the ensemble models is improved with these additional features.
Flow: optical flow features. DCNN: the entire VGG$16$ model fine-tuned by data collected by \CFM3
}
\label{tab:Improving the decision model}
\begin{tabular}{l c}
\hline
& \\ [-2ex]
Method & Avg. miss rate (\%) \\
\hline\hline
& \\ [-2ex]
\CFM3 only        							& $13.49$ \\
\hline
& \\ [-2ex]
\CFM3$+$ACF       							& $12.38$ \\
\hline
& \\ [-2ex]
\CFM3$+$ACF$+$Flow 							& $11.11$ \\
\hline
& \\ [-2ex]
(\CFM3$+$ACF)$+$\CFM4$+$\CFM5$+$DCNN 		& $9.37$ \\
\hline
& \\ [-2ex]
(\CFM3$+$ACF$+$Flow)$+$\CFM4$+$\CFM5$+$DCNN & {$\bf 9.32$} \\
\hline
\end{tabular}
\end{center}
\vspace{-20pt}
\end{table}

Table~\ref{tab:Improving the decision model} shows the detection results of different variants of
\CFM3 model. With adding the ACF features, the MR of \CFM3
detector is reduce by $1.11\%$. With the extra optical flow features,
the MR is further reduced to $11.11\%$.
These experimental results demonstrate that hand-crafted features carry complimentary information
which can further improve the DCNN convolutional features.
This is easy to understand: the ACF features may be viewed as lower-level features, compared with
the middle-level features in \CFM3. The optical flow clearly encodes motion information which is not in \CFM3 features.
By adding the other components of the proposed ensemble model, our detector can achieve $9.32\%$ MR. The MR is slightly increased to $9.37\%$ by removing motion information.

\subsection{Pixel Labelling}

As shown in Section~\ref{sec:approach2}, the pixel labelling model is also complementary to convolutional features.
Table~\ref{tab:pixel labelling} shows the detection performance of different ensemble models enhanced by pixel labelling model.
The best result is achieved by combining the most number of different types of models (which is refer to as All-in-one), which 
reduces the MR on the Caltech $\sf Reasonable$ test set from the previous best $11.7\%$ to $8.9\%$.
Note that the combination rule used by our methods is simple, which implies a potential for further improvement.

\begin{table}[tbp!]
\begin{center}
\caption{Comparison of detection performance (on $\sf Reasonable$) of different ensemble models with pixel labelling.
         DCNN: the entire VGG$16$ model fine-tuned by hard negative data collected by \CFM3;
         Pixel label.: pixel labelling model; Flow: optical flow.
         The pixel labelling model consistently improves all the considered models in this table.
         The All-in-one model set a new record on the Caltech pedestrian benchmark
         }
\label{tab:pixel labelling}
\begin{tabular}{l c}
\hline
& \\ [-2ex]
Method  & Avg. miss rate (\%) \\
\hline\hline
& \\ [-2ex]
\CFM3$+$Pixel label. & $11.58$ \\
\hline
& \\ [-2ex]
\CFM3$+$\CFM4$+$\CFM5$+$Pixel label. &  $9.94$
  \\
\hline
& \\ [-2ex]
\CFM3$+$\CFM4$+$\CFM5$ + $DCNN$ + $Pixel label. &  $9.53$ \\
\hline
& \\ [-2ex]
(\CFM3$+$ACF)$+$\CFM4$+$\CFM5$+$  		& $9.06$ \\
 $\,\,\,$DCNN$+$Pixel label. & \\
\hline
& \\ [-2ex]
(\CFM3$+$ACF$+$Flow)$+$\CFM4$+$\CFM5$+$  & {$\bf 8.93$} \\
 $\,\,\,$DCNN$+$Pixel label. (All-in-one)& \\
 \hline
\end{tabular}
\end{center}
\vspace{-10pt}
\end{table}

\subsection{Ablation Studies}

\begin{table*}[tbp!]
\begin{center}
\caption{Ablation studies of the All-in-one model on the Caltech $\sf{Reasonable}$ test set}
\label{tab:ablation studies}
\scalebox{0.8}
{\footnotesize
\begin{tabular}{l ccccccc}
\hline
& & & & & & & \\ [-2ex]
\multirow{2}{*}{Model} & \CFM3a & \CFM3 & \CFM3$+$\CFM4 & \CFM3$+$\CFM4 & \CFM3$+$\CFM4 & \CFM3$+$\CFM4$+$\CFM5 & All-in-one \\
& & & & $+$\CFM5 & $+$\CFM5$+$DCNN & $+$DCNN$+$Label. &  \\
\hline
& & & & & & & \\ [-2ex]
Pipeline & \CFM3a & fine-tuning & Add \CFM4 & Add \CFM5 & Add DCNN  & Add Pixel Label. & Use (\CFM3+ACF+Flow) \\
\hline\hline
& & & & & & & \\ [-2ex]
Miss rate (\%) & $18.71$ & $13.49$ & $10.68$ & $10.46$ & $10.07$ & $9.53$ & $8.93$ \\
\hline
& & & & & & & \\ [-2ex]
Improve. (\%) & $-$ & $+5.22$ & $+2.81$ & $+0.22$ & $+0.39$ & $+0.54$ & $+0.6$ \\
\hline
\end{tabular}
}
\end{center}
\vspace{-15pt}
\end{table*}

We investigate the overall pipeline of the All-in-one model by adding each
component step by step, which is shown in Table~\ref{tab:ablation studies}. 
As the start point, the \CFM3a model with the original VGG$16$ model pre-trained on ImageNet data achieves
a miss rate of $18.71\%$.  A $5.22\%$  performance gain can be obtained by fine-tuning
the VGG$16$ model with bootstrapped data. The detection
results can be improved to $10.46\%$ (better than all previous methods) by adding \CFM4 and \CFM5 models to
construct an ensemble model. We obtain  $0.39\%$ performance improvement if we use
the entire VGG$16$ model (fine-tuned by bootstrapped data with \CFM3) as a component of our ensemble model.
Combining the pixel labelling information to predicted bounding boxes can
further reduce the miss rate by $0.54\%$. By replacing the \CFM3 model to \CFM3$ + $ACF$ + $Flow model, the 
MR of our ensemble mode can eventually achieve $8.93\%$ on the Caltech $\sf Reasonable$ test set.

\subsection{Fast Ensemble Models}

In this section, we investigate the speed issue of the proposed detector. Our All-in-one model takes about 8s for processing one $640\times480$ image on a workstation with one octa-core Intel Xeon 2.30GHz processor and one Nvidia Tesla K40c GPU. Most of time (about 7s) is spent on the extraction of the CFMs on a multi-scale image pyramid. The remaining components of the ensemble model takes less than 1s to process the passed region proposals. The pixel labelling model only uses about 0.25s to process one image since it only need to be applied to one scale. It can be easily observed that the current bottleneck of the proposed detector is the \CFM3 which is used to extract region proposals with associated detection scores. The speed of our detector can be accelerated using a light-weight proposal method at the start of the pipeline in Fig.~\ref{fig:flow1}. 

We use two pedestrian detectors ACF~\cite{dollar2014fast} and checkerboards~\cite{zhang2015filtered} as the proposal methods. Our ACF detector consists of 4096 depth-4 decision trees, trained via real-Adaboost. The model has size $128 \times 64$ pixels, and is trained via four rounds of bootstrapping. The sliding window stride is 4 pixels. The checkerboards detector is trained using almost identical parameters as for ACF. The only difference is that the feature channels are the results of convolving the ACF channels with a set of checkerboard filters. In our implementation, we adopt a set of 12 binary $2 \times 2$ filters to generate checkerboard feature channels. To limit the number of region proposals, we set the threshold of the above two detectors to generate about 20 proposals per image.

Table~\ref{tab:fast detector} shows the detection performance of the original ensemble model and fast ensemble models on Caltech $\sf Reasonable$ test set. We can observe that the quality of proposals are enhanced by a large margin using both ensemble models and the pixel labelling model. The best result of fast ensemble models is achieved by using proposals generated by the checkerboards detector. This method uses the data collected by checkerboards detector as the initial fine-tuning data. With a negotiable performance loss (e.g., 1.12\%), it's about 6 times faster than the original method. Note that the fast ensemble model (with checkerboard proposals) also achieves the state-of-the-art results.

\begin{table}[tbp!]
\begin{center}
\caption{Comparison of detection performance (on $\sf Reasonable$) between the original ensemble model and fast ensemble models}
\label{tab:fast detector}
\scalebox{0.9}
{\footnotesize
\begin{tabular}{lcc}
\hline
Method & Avg.\ miss rate (\%) & runtime (s)\\
\hline\hline
& \\ [-2ex]
\CFM3 (proposals)$ + $\CFM4$ + $\CFM5$ + $DCNN$ + $Pixel label. & $9.53$ & 8\\
\hline
& \\ [-2ex]
ACF (proposals)$ + $\CFM3$ + $\CFM4$ + $\CFM5$ + $DCNN$ + $Pixel label. & $12.20$ & 0.75\\
\hline
& \\ [-2ex]
Checkerboards (proposals)$ + $\CFM3$ + $\CFM4$ + $\CFM5$ + $DCNN$ + $Pixel label. & $10.65$ & 1.25\\
\hline
\end{tabular}
}
\end{center}
\vspace{-20pt}
\end{table}

\subsection{Comparison to State-of-the-art Approaches}

We compare the detection performance of our models with existing state-of-the-art approaches on the Caltech dataset. 
Table~\ref{tab:comparison of DCNN based methods} compares our models with a wide range of detectors, including 
boosted decision trees trained on hand-crafted features, RCNN-based methods and the state-of-the-art methods on the Caltech $\sf{Reasonable}$ test set.
The performance of the first two types are quite close to each other.
Using only one single layer of convolutional feature map, 
our \CFM3 model has outperformed all other methods expect the two sophisticated methods~\cite{tiandeep,cai2015learning}.
Note that the RCNN based methods are based on larger models than \CFM3.
As feature representation, the CFM from the Conv$3$-$3$ layer of our fine-tuned model significantly outperforms all other hand-crafted features. The \CFM3$+$Pixel labelling model is comparable to the state-of-the-art performance achieved by sophisticated methods~\cite{tiandeep,cai2015learning}. Our \CFM3$+$\CFM4$+$\CFM5 model performs even better. Without using hand-crafted features, our model can achieve $9.53\%$ MR. The best result is achieved by the All-in-one model which combines a number of hand-crafted features and \CFM{} models.

\begin{table}[tbp!]
\begin{center}
\caption{Detection performance of different types of detectors on the Caltech 	
		$\sf{Reasonable}$ test set.
         Three types of approaches are compared in this table, including boosted decision 
         trees trained on hand-crafted features, 
         RCNN-based methods and the state-of-the-art sophisticated methods.
         All of our models outperform the first three types of models, and our All-in-one set a new recorded MR on Caltech 	
         pedestrian benchmark. $^\dagger$ indicates the methods trained with optical flow features
}
\label{tab:comparison of DCNN based methods}
\scalebox{0.9}
{\footnotesize
\begin{tabular}{llc}
\hline
& &  \\ [-2ex]
Type & Method & Miss Rate (\%) \\
& &  \\ [-2ex]
\hline\hline
& &  \\ [-2ex]
\multirow{1}{*}{Hand-crafted Features}
& SpatialPooling~\cite{paisitkriangkrai2014strengthening} & $29.24$\\
& SpatialPooling$+$~\cite{paisitkriangkrai2014pedestrian}$^\dagger$ & $21.89$\\
& LDCF~\cite{nam2014local} & $24.80$\\
& Checkerboards~\cite{zhang2015filtered} & $18.47$\\
& Checkerboards$+$~\cite{zhang2015filtered}$^\dagger$ & $17.10$\\
& &  \\ [-2ex]
\hline
& &  \\ [-2ex]
\multirow{1}{*}{RCNN based}
& AlexNet~\cite{hosang2015taking}   & $23.32$ \\
& GoogLeNet~\cite{tiandeep} & $16.43$ \\
& &  \\ [-2ex]
\hline
& &  \\ [-2ex]
\multirow{1}{*}{State-of-the-arts}
&DeepParts~\cite{tiandeep} & $11.89$ \\
&CompACT-Deep~\cite{cai2015learning} & $11.75$ \\
& &  \\ [-2ex]
\hline
& &  \\ [-2ex]
\multirow{1}{*}{Ours}
&\CFM3                 & ${13.49}$ \\
&\CFM3$+$Label.		   & ${11.58}$ \\
&\CFM3$+$\CFM4$+$\CFM5 & ${10.46}$ \\
&\CFM3$+$\CFM4$+$\CFM5+DCNN+Label. & ${9.53}$ \\
&All-in-one$^\dagger$  & $\textbf{8.93}$ \\
& &  \\ [-2ex]
\hline
\end{tabular}
}
\end{center}
\vspace{-10pt}
\end{table}

\begin{figure}[tbp!]
\begin{center}
	\includegraphics[width=0.8\textwidth]{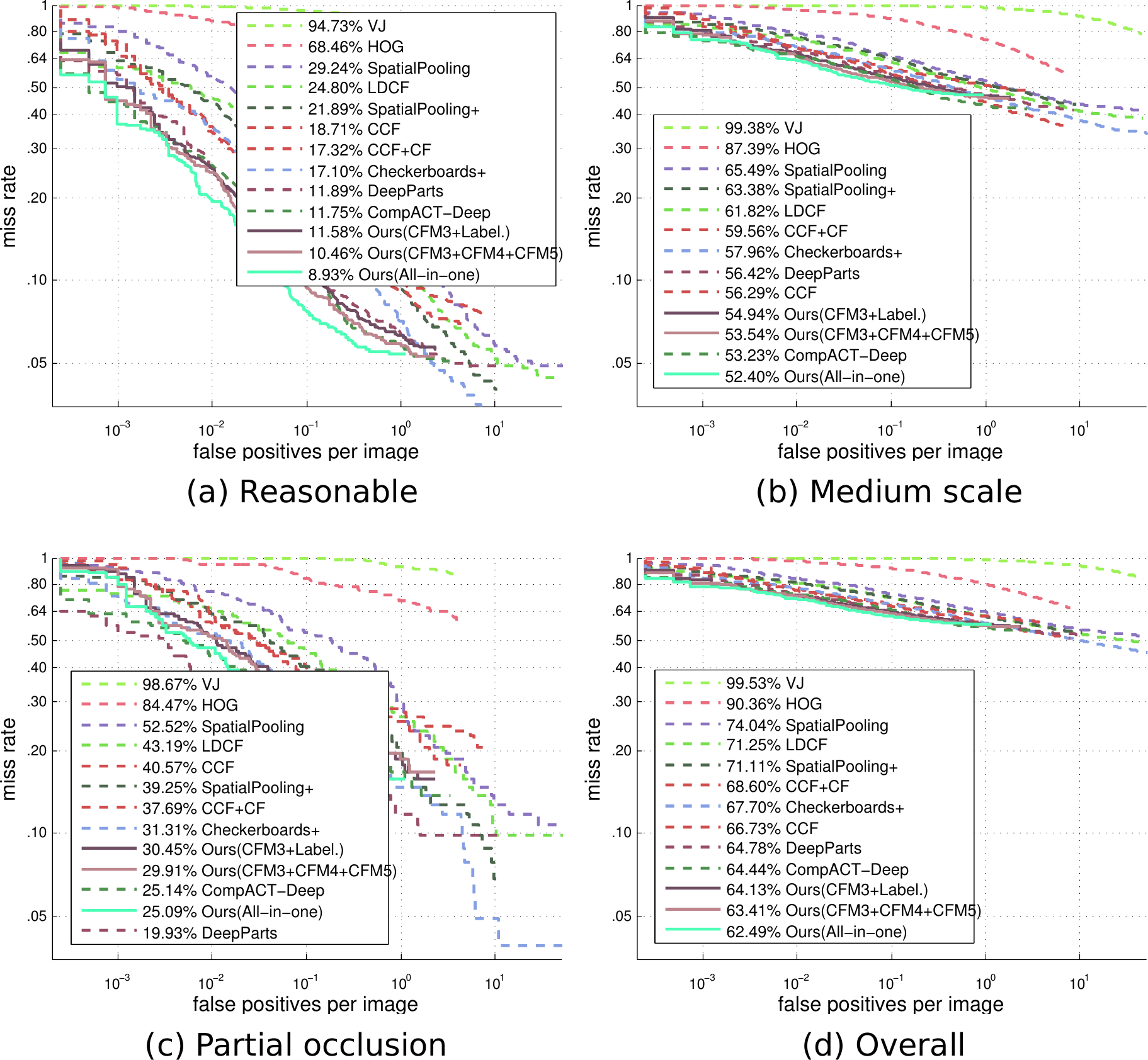}
\end{center}
\caption{Comparison to state-of-the-art on various Caltech test settings.}
\label{fig:caltech_evaluation}
\end{figure}

Fig.~\ref{fig:caltech_evaluation} shows a more complete evaluation of the proposed detection framework on various Caltech test settings, including $\sf{Reasonable}$, $\sf{Medium~scale}$, $\sf{Partial~occlusion}$, and $\sf{Overall}$. We can observe that our ensemble model achieves the best results on most test subsets (including $\sf{Reasonable}$). On the $\sf{Partial~occlusion}$ set, our models are only outperformed by DeepParts~\cite{tiandeep}, which is specifically trained for handling occlusions.

\begin{minipage}[tbp!]{\textwidth}
\begin{minipage}[l]{0.45\textwidth}
\centering
\captionof{table}{Detection results on the KITTI dataset. Note: $^\ast$ indicates the methods trained with stereo images}
\label{tab:comparison on kitti}
\scalebox{0.8}
{\footnotesize
\begin{tabular}{c | c c c}
\hline
Method & Moderate(\%) & Easy(\%) & Hard(\%) \\
\hline\hline
& & & \\ [-2ex]
3DOP$^\ast$~\cite{chen20153d} & $67.47$ & $81.78$ & $64.70$ \\
Fast-CFMs (Ours) & $63.26$ & $74.22$ & $56.44$ \\
Reionlets~\cite{wang2013regionlets} & $61.15$ & $73.14$ & $55.21$ \\
CompACT-Deep~\cite{cai2015learning} & $58.74$ & $70.69$ & $52.71$ \\
DeepParts~\cite{tiandeep} & $58.67$ & $70.49$ & $52.78$ \\
FilteredICF~\cite{zhang2015filtered} & $56.75$ & $67.65$ & $51.12$ \\
pAUCEnsT~\cite{paisitkriangkrai2014pedestrian} & $54.49$ & $65.26$ & $48.60$ \\
R-CNN~\cite{hosang2015taking} & $50.13$ & $61.61$ & $44.79$ \\
\hline
\end{tabular}
}
\vspace{-20pt}
\end{minipage}
\hfill
\begin{minipage}[c]{0.5\textwidth}
\centering
	\includegraphics[width=0.8\textwidth]{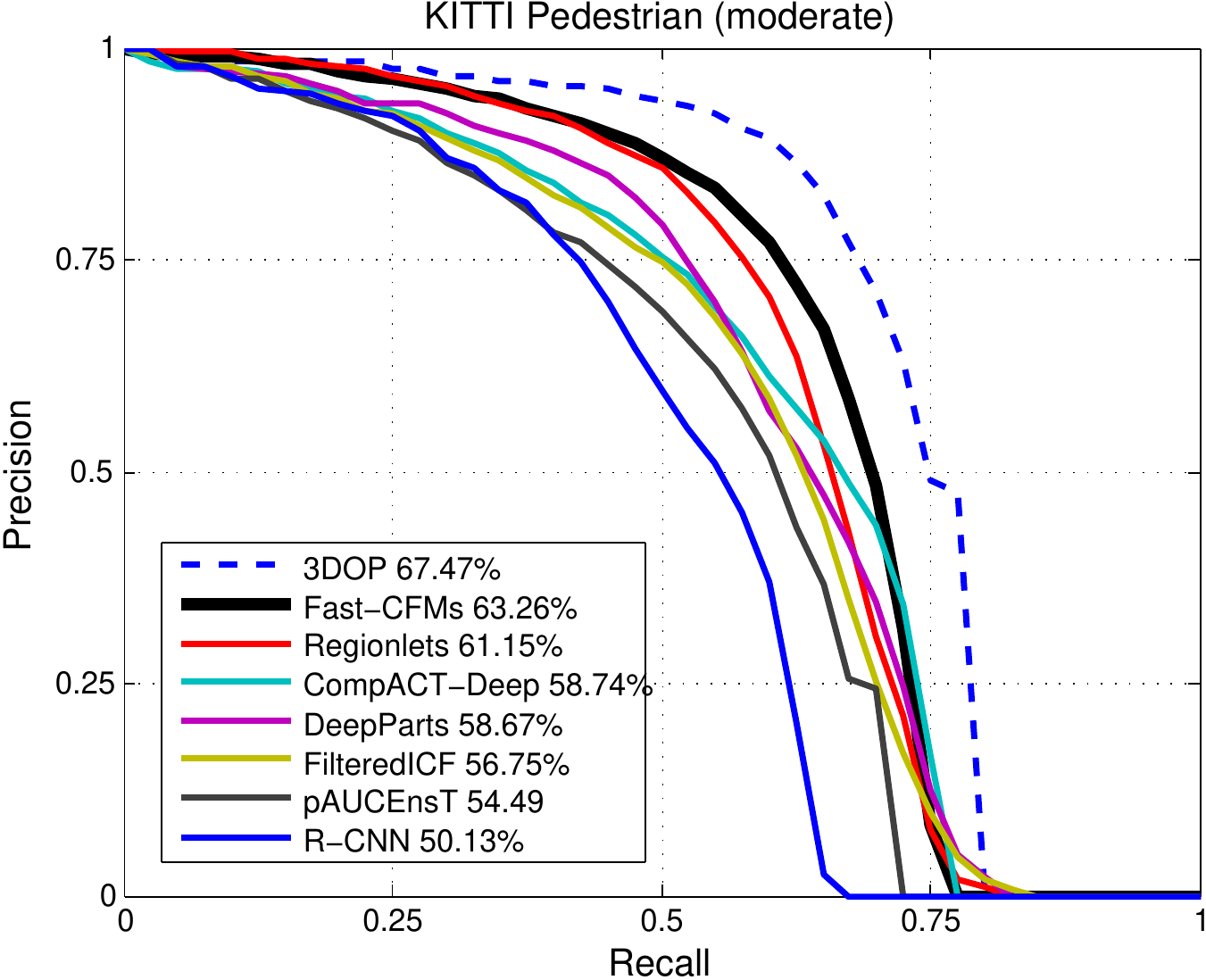}
\captionof{figure}{Comparison to state-of-the-art on the KITTI $\sf{Moderate}$ test set.
}
\vspace{-50pt}
\label{fig:kitti_moderate}
\end{minipage}
\vspace{+50pt}
\end{minipage}

Table~\ref{tab:comparison on kitti} shows the results on the KITTI dataset. Since images of KITTI are larger than in Caltech, the feature extraction of \CFM3{} model is time-consuming. In our experiments, only the fast ensemble model with Checkerboards proposals %
is used for testing on KITTI. Our model achieves competitive results, $74.22\%$, $63.26\%$, and $56.44\%$ AP on $\sf{Easy}$, $\sf{Moderate}$, and $\sf{Hard}$ subsets respectively. Fig.~\ref{fig:kitti_moderate} presents the comparison of detection performance on the KITTI $\sf{Moderate}$ test subset. It can be observed that the proposed detector outperforms all published monocular-based methods. Note that the 3DOP~\cite{chen20153d} is based on stereo images. The proposed ensemble model is the best-performing detector based on DCNN, and surpasses CompACT-Deep~\cite{cai2015learning} and DeepParts~\cite{tiandeep} by $4.52\%$ and $4.59\%$ respectively.

\section{Conclusions}

In this work, we have built a simple-yet-powerful pedestrian detector, which re-uses
inner layers of convolutional features extracted by a properly fine-tuned VGG16 model.
This `vanilla' model has already achieved the best reported results on the Caltech dataset, using the same training data as
previous DCNN approaches. With a few simple modifications, its variants have achieved even more significant results.

We have presented extensive and systematic empirical evaluations on the
effectiveness of DCNN features for pedestrian detection.
We show that it is possible to build the best pedestrian detector, yet
avoiding complex custom designs. We also show that a pixel labelling model can be used to
improve performance by simply incorporating the labelling scores with the detection scores
of a standard pedestrian detector. Note that simple combination rules are used here, which leaves potentials for further improvement. For example the ROI pooling for further speed and performance improvement.

\bibliographystyle{splncs}
\bibliography{egbib}

\end{document}